\documentclass[letterpaper, 10 pt, conference]{ieeeconf}
\usepackage{cite}
\IEEEoverridecommandlockouts    

\usepackage{amsmath}
\usepackage{amssymb}
\usepackage{bm}
\usepackage{booktabs}
\usepackage{cite}
\usepackage{graphicx}
\usepackage{lettrine}
\usepackage{multirow}
\usepackage{subcaption}
\usepackage[table]{xcolor}
\usepackage{xspace}
\usepackage[normalem]{ulem}
\useunder{\uline}{\ul}{}
\usepackage{hyperref}

\newcommand{\method}{\textit{FleetAgent}\xspace}

\begin{document}
%
\title{FleetAgent: Teleoperation Assistant for Autonomous Fleets via\\Vectorized V2N Messages}

%
%

\author{
Juntong Peng$^{1}$, Qi Chen$^{2}$, Deyuan Qu$^{2}$, Takayuki Shimizu$^{2}$, Yaobin Chen$^{1}$, and Ziran Wang$^{1}$%
\thanks{$^{1}$Juntong Peng, Yaobin Chen, and Ziran Wang are with the College of Engineering, Purdue University, West Lafayette, IN, USA. Corresponding author: Juntong Peng~{\tt\small juntong@purdue.edu}.}%
\thanks{$^{2}$Qi Chen, Deyuan Qu, and Takayuki Shimizu are with Toyota InfoTech Labs, Mountain View, CA, USA.}%
}

%
%


\maketitle

\begin{abstract}
Large-scale autonomous fleets rely on teleoperation to resolve rare failures, yet streaming raw sensor data from many vehicles is costly, and remote operators can only monitor a limited number of vehicles at a time.
We introduce \method, a cloud-hosted multimodal large language model (MLLM) assistant that consumes compact vectorized vehicle-to-network (V2N) messages, such as map elements, detected objects, and the ego planned path. It provides a structured natural-language response (including narration, explanation, and evaluation of the plan and scene), along with an intervention urgency score for operator prioritization.
To make structured messages compatible with token-based MLLMs, we propose \textbf{VecFormer}, a vector-to-embedding interface with differentiable top-$K$ context selection that bounds context length and GPU KV-cache growth, enabling more efficient batch processing, which is important under the context of cloud-hosted large-scale fleet management.
We also construct \textbf{VecEval}, a nuScenes-derived dataset with paired human and synthetic imperfect plans and human-verified language labels, to facilitate the training and evaluation of our proposed system.
Our proposed system can reduce uplink payload by up to $625\times$ compared with raw images and reduce KV-cache memory by $16.54\times$ compared with original text descriptions.
On VecEval, \method improves Lingo-Judge score by $16.8$\% and reduces intervention failure rate by $19.9$\%, compared with Qwen2.5-VL-7B using language descriptions.
These results demonstrate that \method can utilize compact structured V2N messaging to enable efficient, explainable teleoperation monitoring for autonomous fleets.
\end{abstract}


\section{Introduction}


Autonomous driving technology has been evolving rapidly over the last decade, with companies like Waymo, Zoox, and Tesla starting to deploy driverless fleets to the general public in major U.S.\ cities \cite{waymo, zoox, tesla_robotaxi}.
Large-scale driverless fleets sometimes rely on teleoperation as a backup to handle rare and safety-critical situations that the onboard system fails to resolve.
Remote operation can take the forms of remote monitoring, direct control, or human guidance.
In this work, we focus on the \textbf{remote monitoring and operator prioritization} setting: continuously identifying vehicles that are likely to require operator attention, and generating concise, structured explanations that help operators quickly rebuild \textbf{situational awareness} (scene context, ego intent, key interactions, and reasons for potential intervention) when switching between vehicles.

Typical remote driving systems feature bi-directional communication, where an autonomous vehicle's sensor data, such as camera and LiDAR feeds, is transmitted over a wireless network to a human operator at a remote center.
The operator assesses the situation and sends guidance back to the vehicle.
To mitigate safety risks associated with latency, some industry leaders adopt shared-autonomy designs where the autonomous vehicle (AV) remains in control of low-level operations while the human provides high-level guidance.
However, vehicle-to-server sensor streaming remains challenging regarding latency, stability, and cost as the system scales up.


Beyond bandwidth constraint, a practical bottleneck is the operator's \emph{context-switch cost}: when attention shifts from one vehicle to another, the operator must rapidly reconstruct what is happening. This motivates an intelligent assistant that continuously evaluates vehicle behavior and, at the right moments, alerts operators with a structured narration, explanation, and evaluation, while transmitting only a compact vehicle-to-network (V2N) message.

\begin{figure*}[t]
    \centering
    \includegraphics[width=\linewidth]{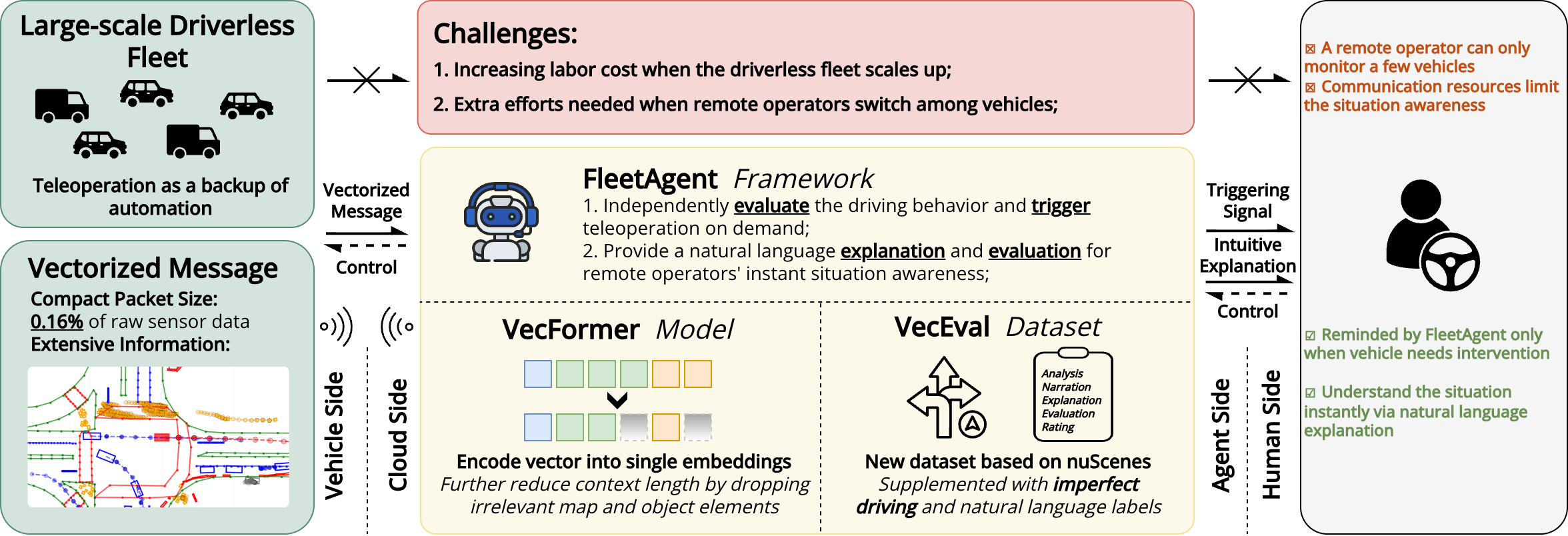}
    \caption{\method: A teleoperation assistant framework for large-scale driverless fleet, providing intervention urgency rating and natural language explanations to support operator situational awareness.}
    \label{fig:abstract}
\end{figure*}

Multi-modal large language models (MLLMs) have shown reasoning capabilities for complex tasks, including autonomous driving.
This makes them promising for identifying challenging scenarios and questionable driving decisions across large fleets, while naturally producing operator-facing explanations that accelerate situational awareness.
However, existing driving-related MLLM pipelines typically assume access to raw sensor inputs, which are too large and costly to transmit over V2N links.

In this paper, we present \method, an on-cloud teleoperation assistant that evaluates ego planning using compact vectorized V2N messages and provides structured natural-language responses plus an urgency score to support operator prioritization. Contributions are summarized as follows:
\begin{itemize}
    \item \textbf{Fleet-scale teleoperation assistant formulation and pipeline:} We propose \method, which evaluates a vehicle's planned behavior from compact vectorized V2N messages and produces (a) structured narration, explanation, and evaluation for rapid \emph{operator situational awareness}; and (b) an intervention urgency score for \emph{operator prioritization}.
    \item \textbf{VecFormer interface for bounded MLLM inference:} We introduce \textit{VecFormer}, a vector-to-embedding interface with differentiable top-$K$ context selection, enabling bounded-length multimodal input tokens and substantially reducing KV-cache growth compared with tokenized text descriptions.
    \item \textbf{VecEval dataset for plan evaluation and operator prioritization:} We construct \textit{VecEval} on top of nuScenes \cite{caesar_nuscenes_2020}, providing paired human and imperfect plan variants with human-verified structured language labels and intervention urgency scores.
    \item \textbf{System- and model-level evaluation:} We quantify bandwidth, latency, and memory footprint improvements and demonstrate that \method maintains competitive plan evaluation performance under strict communication and computation budgets.\footnote{The code will be available at:
\url{https://github.com/JuntongPeng/FleetAgent}.}
\end{itemize}

\section{Related Work}

\subsection{Remote Driving Systems}
Prior work on vehicular and robotic teleoperation has primarily studied real-time control under network latency and bandwidth constraints.
Georg et al. quantify latency effects~\cite{georg_sensor_2020}, and Neumeier et al. propose a data-rate reduction method for teleoperated driving video streams~\cite{neumeier_data_2022}.
Shared-autonomy formulations combine human high-level input with vehicle low-level planning~\cite{schitz_shared_2021}, and predictive displays reduce perceived latency by visualizing near-future system states~\cite{xie_generative_2021}.
Other work addresses operational aspects such as selecting suitable remote operators~\cite{gohar_cost_2020} and improving teleoperation interfaces~\cite{wolf_should_2024, tsagkournis_supervised_2023, yang_assisted_2020,cho_goondae_2023}.

In contrast to control or interface-centric teleoperation, our focus is on fleet-scale operator prioritization: identifying which vehicles are likely to require operator attention and providing fast, interpretable situational awareness from compact V2N messages.
This complements edge-case detection and ODD-based triggers surveyed in \cite{rahmani_systematic_2024} by providing teleoperator-facing explanation and evaluation of the planned behavior.

\subsection{MLLMs in Autonomous Driving}
The application of MLLMs in the autonomous driving domain has evolved into multiple paradigms~\cite{wang_generative_2025,sima_drivelm_2025,tian_drivevlm_2024,park_nuplanqa_2025}, including visual question answering (VQA) and integrated end-to-end systems~\cite {li_recogdrive_2025,hwang_emma_2024,zhou_opendrivevla_2025,xu_drivegpt4_2024}, where multimodal LLMs are used for the AV planning. Agent-Driver~\cite{mao_language_2024} proposed an LLM-agent-based method to provide explainable actions. To enhance the aligned interpretability, Hint-AD~\cite{ding_hint-ad_2024} generates language output aligned with the autonomous driving model output, and the paper provided a driving explanation dataset Nu-X. Moreover, ALN-P3~\cite{ma_aln-p3_2025} proposed a distillation approach to transfer the knowledge from multimodal LLM to a light autonomous driving model. Notably, in most prior research, due to the actual needs of the VQA task and vehicle planning task, multimodal LLMs are designed to be deployed onboard, and the model inputs typically include language instructions, raw sensor data, and ego state information. This information can comprehensively describe what's happening around and inside the vehicle, but is too large to be transmitted over the network.

\subsection{Vehicle-to-everything Communication}

Vehicle-to-everything (V2X) communication, including DSRC and C-V2X, enables network-based data exchange between vehicles and infrastructure \cite{abboud2016interworking}.
V2N connects vehicles to cloud infrastructure and is particularly used for teleoperation.
Even with modern 5G links, sensor streaming can remain high-latency and unstable \cite{testouri_5g-enabled_2025}, and reducing payload size can significantly improve transmission under congestion \cite{zhao_quantv2x_2025}.
These observations motivate our focus on compact, vectorized V2N messages for scalable fleet operation and improve the system's feasibility under constraints and a congested network environment.

\section{Problem Formulation}\label{sec:formulation}

\begin{figure*}[t]
    \centering
    \includegraphics[width=\linewidth]{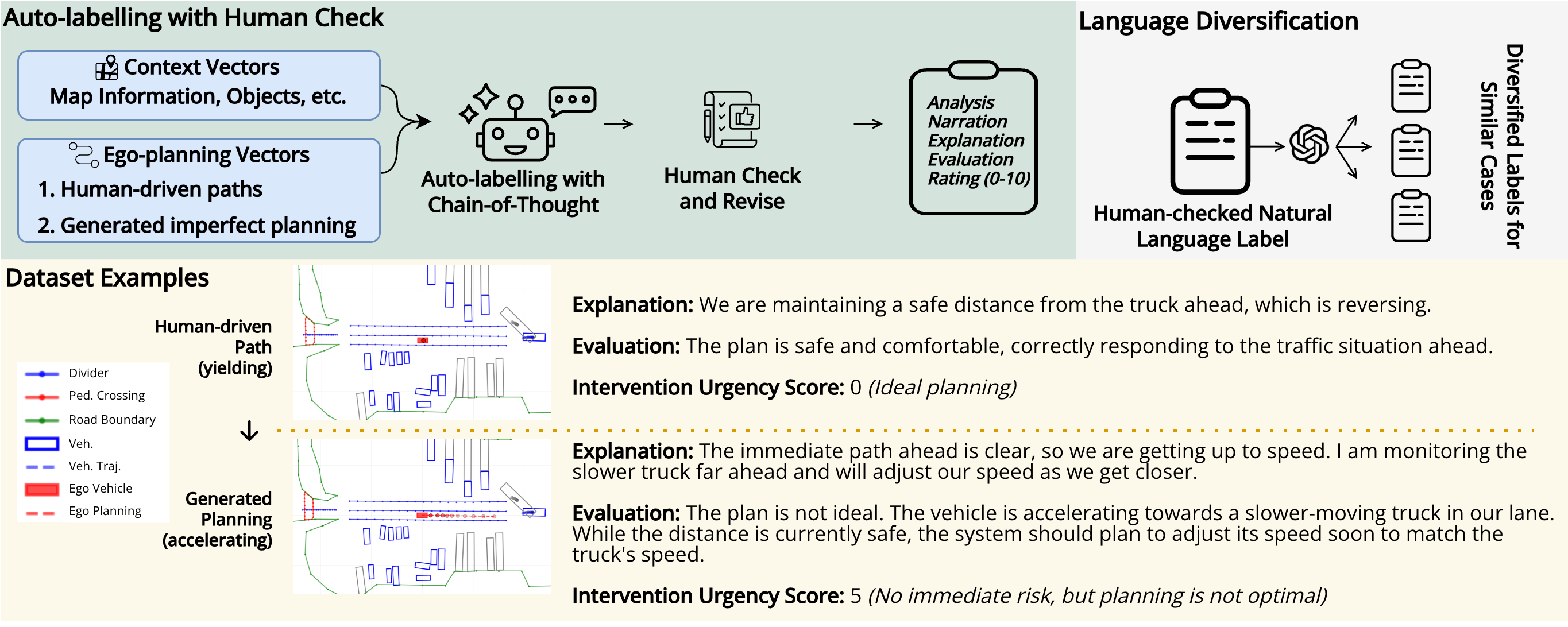}
    \caption{VecEval annotation pipeline and examples. We pair a human-driven plan from the original dataset with generated imperfect plan variants and annotate structured language responses with an intervention urgency score (0--10).}
    \label{fig:ann}
\end{figure*}

We formulate teleoperation assistance as conditional text generation and scoring based on a vehicle's planned motion and its surrounding context.
Traditional driving VQA often follows $f(I,Q)\!\to\!A$, where raw sensor input $I$ and a question $Q$ produce a natural language answer $A$.
End-to-end planning models map $f(I,S)\!\to\!P$ from sensor input $I$ and ego state $S$ to an ego planned trajectory $P$.
In contrast, we consider the explanation and evaluation setting on the cloud:
\begin{equation}
    f(M, P) \rightarrow (L, i),
\end{equation}
where $M$ is a structured message derived from the observation of the scene (map elements, dynamic objects, and their predicted motion) and $P$ is the ego planned trajectory, both produced by onboard autonomy stack.
The outputs include a structured natural-language response $L$ intended to support operator situational awareness by summarizing the scene, ego intent, key interactions, and risk factors, and an intervention urgency score $i \in [0,10]$ used for operator prioritization. Any operator guidance is then sent through the existing teleoperation system outside \method.

This formulation introduces challenges on communication constraints through V2N, and computational constraints on processing large-scale data from fleets using MLLM.
\method addresses these challenges by using compact vectorized V2N messages and a dedicated vector-to-embedding interface (VecFormer) for efficient MLLM inference.
\section{Dataset}

We construct \textit{VecEval}, a dataset for teleoperation assistance that pairs structured driving context with multiple candidate plans and provides human-verified language-based plan evaluation along with an intervention urgency score.
This section summarizes why existing datasets are insufficient for our setting (Sec.~\ref{sec:formulation}) and details the construction pipeline.

\subsection{Task Requirement and Dataset Positioning}

Our task requires evaluating a given ego plan $P$ under a structured observation $M$ (e.g., lanes, boundaries, nearby agents with predicted motion).
Each training instance should provide those structured observations and ego vehicle planning results, used as the model input. The annotation of each instance should include natural language labels that explicitly assess plan quality and safety, along with the corresponding intervention urgency score.

Large-scale driving datasets such as nuScenes~\cite{caesar_nuscenes_2020}, Waymo Open~\cite{sun_scalability_2020}, KITTI~\cite{geiger_vision_2013}, and nuPlan~\cite{caesar_nuplan_2022} provide rich sensor logs and high-quality trajectories driven by human drivers.
These datasets supply the context $M$ and an expert plan candidate, but they do not provide enough imperfect plan variants that represent plausible but unsafe or inefficient plans. Structured language evaluation and urgency labels specifically needed for our task are also not included.

Several nuScenes or nuPlan extensions target captioning, grounding, or scene understanding (e.g., NuScenes-QA~\cite{qian_nuscenes-qa_2024}, NuInstruct~\cite{Ding_2024_CVPR}, NuPrompt~\cite{wu_language_2025}, NuPlanQA~\cite{park_nuplanqa_2025}, DriveLM~\cite{sima_drivelm_2025}).
Driving explanation datasets such as Nu-X~\cite{ding_hint-ad_2024} and BDD-X~\cite{ferrari_textual_2018} provide language labels aligned with human actions, but they do not provide evaluation of alternative candidate plans nor an intervention urgency score targeted at teleoperation management.
VecEval is designed to fill this gap by providing both safe and imperfect plans with a structured language label and a graded urgency signal.

\subsection{Dataset Construction Pipeline}

As shown in Fig.~\ref{fig:ann}, we construct VecEval in two stages: generating imperfect plan variants and annotating structured language responses with urgency scores.

\subsubsection{Imperfect Plan Generation}
Real-world autonomous systems can produce unsafe or inefficient plans due to upstream errors or decision-making failures.
To generate large-scale, kinematically feasible imperfect plans without requiring massive failure logs, we sample counterfactual trajectories from a real trajectory vocabulary derived from human driving.
Concretely, we collect human-driven trajectories from nuScenes and cluster them into 15 categories using the K-Means algorithm, capturing coarse maneuver and speed-profile differences (e.g., straight or turning with acceleration or deceleration). This categorization covers a broad range of common driving behaviors.
For each scenario in the dataset, we generate a counterfactual candidate plan by selecting a category different from the human-driven plan and sampling a trajectory from that category.
This approach produces physically plausible but behaviorally imperfect plans. For example, switching from maintaining a stable speed to accelerating may cause a collision with the front vehicle, while the maneuver is feasible with a mistakenly planned target speed. Other cross-category substitutions can lead to hazards, including lane or road boundary violations.

\subsubsection{Language and Urgency Annotation}

Given a context $M$ and a plan $P$ (human-driven or imperfect), we annotate a structured response consisting of narration, explanation, evaluation, and an intervention urgency score $i\in[0,10]$.
We adopt a two-step process:
(1) Auto-labeling: we use a strong reasoning model (Gemini-2.5-Flash Thinking) with a structured prompt to draft responses based on nuScenes ground truth (map, agents, and plan). Chain-of-Thought prompting is used in this process to improve the draft quality.
(2) Human verification: human annotators review and edit the drafts to ensure factual correctness, remove hallucinations, and calibrate urgency scores according to the scenario.
Only the verified responses are used for training and evaluation.
The urgency score is defined by three well-defined anchor scores: 0 for safe and efficient maneuvers, 5 for safe but inefficient maneuvers, and 10 for immediate safety issues. LLMs and human annotators are asked to follow the rubrics to generate or verify the assigned urgency score.

\subsection{Examples and Key Statistics}

We sample human-driven instances along trajectories to ensure coverage while maintaining label diversity. Specifically, 11,510 out of 28,130 training samples and 1,693 out of 6,019 validation samples from the original nuScenes split are selected, yielding an average spatial interval of $4.43$~m between annotated frames. We additionally annotate imperfect plan variants and the corresponding responses for some scenarios. In total, VecEval contains 12,754 training samples and 1,982 validation samples. Examples are shown in the lower part of Fig.~\ref{fig:ann}.
\section{Methodology}

\subsection{Overall Architecture}

As shown in Fig.~\ref{fig:arch}, \method operates as an on-cloud assistant that monitors driving behavior from compact V2N messages and produces an operator-facing explanation plus an intervention urgency score.
Unlike driving VQA pipelines that input raw images, we use a vectorized representation to balance transmitted information and communication cost.
Road structure, dynamic agents, and ego planning can all be represented as compact vectors that are already available in typical autonomy stacks, introducing no additional onboard computation.

A straightforward way to feed vectors into an LLM is to convert them into a long text description.
While simple, this induces very long contexts (often more than $10{,}000$ tokens per frame), which increases runtime memory and reduces throughput, especially in complex scenes that are more likely to require teleoperation.
To address this, we introduce \textit{VecFormer}, which converts vectorized inputs into bounded-length context tokens compatible with the base MLLM.

We use Qwen2.5-VL \cite{bai_qwen25-vl_2025} as the base model because it supports a unified embedding interface for multimodal fusion, allowing external embeddings to be inserted as ``multimodal tokens'' concatenated with text tokens.

\subsection{Vectorized V2N Message Representation}
Each vehicle transmits a compact message containing:
(i) Map elements (e.g., lane dividers, boundaries, crosswalks) represented as polylines,
(ii) Dynamic objects represented by current position, heading, and predicted future trajectories,
and (iii) Ego plan represented by future waypoints.
We denote map vectors as $\{\mathcal{M}_i\}$, object vectors as $\{\mathcal{O}_j\}$, and the ego plan as $\mathcal{P}$.
This structured input provides the minimum context needed for plan evaluation while requiring significantly less message volume than raw sensor streaming.

\subsection{\textit{VecFormer} Design}

\begin{figure}[t]
    \centering
    \includegraphics[width=\linewidth]{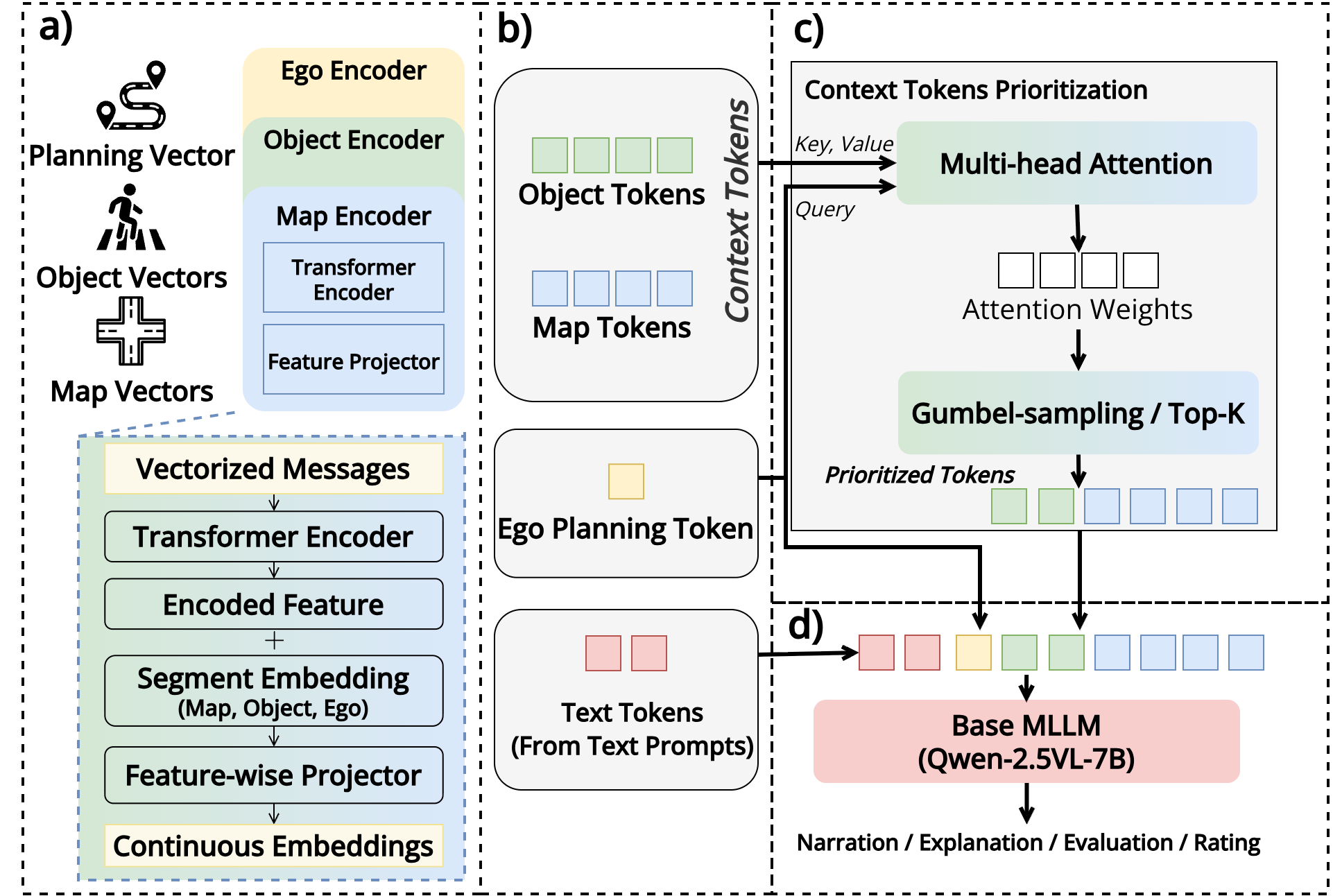}
    \caption{\textit{VecFormer} design. Multi-modal encoders (a) encode vectorized map, object, and ego-plan inputs into LLM-compatible continuous embeddings (b) and apply differentiable top-$K$ context selection (c) to prioritize planning-relevant context. Base MLLM (d) consumes the multi-modal tokens and provides natural language responses.}
    \label{fig:arch}
\end{figure}

\begin{table*}[t]
\centering
\caption{\textbf{System-level comparison.} API results use OpenAI GPT-4o; local results use Qwen-2.5VL-7B. Memory footprint reports KV-cache size / model parameter memory; the multiplier indicates KV-cache size relative to FleetAgent. Response time is measured on a single NVIDIA A100-40GB with batch size 1.}
\label{tab:system}
\resizebox{\linewidth}{!}{%
\begin{tabular}{cccccc}
\hline
 &
  Model Type &
  Input Type &
  \begin{tabular}[c]{@{}c@{}}Transmitted Packet Size (kB)\\ per request\end{tabular} &
  \begin{tabular}[c]{@{}c@{}}Response Time (s)\\ avg. (s.d.)\end{tabular} &
  \begin{tabular}[c]{@{}c@{}}Memory Footprint (MB)\\ Cache/Model\end{tabular} \\ \hline
\rowcolor[HTML]{EFEFEF} 
1 &
  \cellcolor[HTML]{EFEFEF} &
  Raw Images &
  25312.5 &
  5.8783 (1.5160) &
  - \\
\rowcolor[HTML]{EFEFEF} 
2 &
  \cellcolor[HTML]{EFEFEF} &
  BEV Images &
  4218.8 &
  6.9310 (2.7076) &
  - \\
\rowcolor[HTML]{EFEFEF} 
3 &
  \multirow{-3}{*}{\cellcolor[HTML]{EFEFEF}API} &
  Language Description &
  40.5 &
  7.5245 (3.0132) &
  - \\
4 &
   &
  Raw Images &
  \cellcolor[HTML]{FFFFFF}25312.5 &
   12.6158 (1.0635)&
  10558 / 15819 ($\times 8.51$)\\
5 &
   &
  BEV Images &
  \cellcolor[HTML]{FFFFFF}4218.8 &
  5.8987 (1.0530) &
  2604 / 15819 ($\times 2.10$)\\
6 &
  \multirow{-3}{*}{Local} &
  Language Description &
  \cellcolor[HTML]{FFFFFF}40.5 &
  8.6372 (5.5699) &
  20522 / 15819 ($\times 16.54 $)\\ \hline
\rowcolor[HTML]{EFEFEF} 
\textbf{FleetAgent} &
  \textbf{Local} &
  \textbf{Vectorized V2N Messages} &
  \textbf{40.5} &
  \textbf{4.4116 (1.0610)} &
  \textbf{1241 / 17081} \\ \hline
\end{tabular}%
}
\end{table*}

As shown in Fig.~\ref{fig:arch}, VecFormer consists of two learnable modules:
(1) \textbf{vector encoders} that map each input vector (polyline/trajectory) to a single continuous embedding,
and (2) a \textbf{context token prioritization module} that selects the top-$K$ most relevant map/object embeddings conditioned on the ego planning embedding.

Let $\mathcal{F}_{map,i}$ be the embedding of map vector $\mathcal{M}_i$, $\mathcal{F}_{obj,j}$ be the embedding of object vector $\mathcal{O}_j$, and $\mathcal{F}_{ego}$ be the embedding of ego plan $\mathcal{P}$.
VecFormer produces a set of context embeddings $\mathcal{F}_{ctx}=\{\mathcal{F}_{map},\mathcal{F}_{obj}\}\in\mathbb{R}^{n\times d}$ and an ego embedding $\mathcal{F}_{ego}\in\mathbb{R}^{d}$, where $d$ matches the base MLLM embedding dimension.

To prioritize context, we compute multi-head attention scores using the ego plan as query:
\begin{align}
    \alpha_i = \frac{(W_Q \mathcal{F}_{ego})^T (W_K \mathcal{F}_{ctx,i})}{\sqrt{d_{head}}}
\end{align}
where $W_Q,W_K\in\mathbb{R}^{d_{head}\times d}$ are learned projections and $d_{head}$ is a hyperparameter.
We then select exactly $K$ context vectors via sequential Gumbel-Softmax sampling \cite{jang_categorical_2017} to maintain hard-selection while enabling end-to-end training:
\begin{align}
    y_{soft,i}^{(k)} =
    \frac{\exp((\log(\pi_i)+g_i)/\tau)}
    {\sum_{j\in \mathcal{U}^{(k)}} \exp((\log(\pi_j)+g_j)/\tau)}
\end{align}
where $\pi_i=\text{softmax}(\alpha_i)$, $g_i$ is Gumbel noise, $\tau$ is temperature, and $\mathcal{U}^{(k)}$ indexes unselected contexts.
We use the straight-through estimator $y^{(k)}=y_{hard}^{(k)}-\text{detach}(y_{soft}^{(k)})+y_{soft}^{(k)}$ to keep discrete selection with gradient flow.
The selected embeddings are then concatenated with text prompt tokens and fed into the base MLLM.
In this work, $K$ is chosen as a fixed hyperparameter balancing inference cost and context coverage. However, while the cross-attention modules are end-to-end trained, a flexible context budget, such as confidence-based filtering, is a natural extension when scene complexity varies.

In addition to the efficiency brought by compact V2N messages and a dedicated VecFormer, the design is also teleoperator-oriented: it explicitly links the ego plan to the most relevant agents and map elements, helping teleoperators regain situational awareness quickly when switching between vehicles.

\subsection{Training Strategy}
To adapt VecFormer to a pretrained MLLM with limited human-verified labels, we use three stages to gradually improve the model's capability: masked vector reconstruction, fixed-format instruction tuning, and supervised fine-tuning. The multi-stage training pipeline facilitates self-supervised learning and fixed-template generation to reduce the demand on human annotations.

\noindent\textbf{Masked Vector Reconstruction.}
We pretrain the vector encoders by masking points within vectors and reconstructing the original vectors with a lightweight decoder head.
This encourages the encoder to capture spatial and temporal structure (e.g., polyline continuity, motion trends), without requiring extra natural language labels.

\noindent\textbf{Fixed-format Instruction Tuning.}
We align the encoded feature space with the base MLLM embedding space by pairing embeddings with short prompts that ask the model to interpret the input vectors (e.g., describing the ego trajectory.)
By freezing most of the parameters in LLM, this stage aims to align the encoded feature space with the LLM input embeddings. Similar to the first stage, this stage requires no extra natural language annotations.

\noindent\textbf{Supervised Fine-tuning:} In this stage, we use the pretrained weights from previous stages as the initial weights. During training, only a one-layer feature projector and the context tokens prioritization module are unfrozen. The base LLM is finetuned using prefix tuning and LoRA adapter, preserving the capability for general reasoning from the base LLM, as our task requires extensive reasoning capability via the narration, explanation, evaluation, and the intervention urgency scoring pipeline.

\section{Empirical Results}

In this section, we evaluate \method from two perspectives.
\textbf{System-level} results quantify bandwidth demand, end-to-end latency (including inference latency and communication latency), and GPU memory footprint under different input modalities.
\textbf{Model-level} results quantify ego planning evaluation performance and explanation quality on VecEval, supplemented with a cross-dataset validation on Nu-X~\cite{ding_hint-ad_2024}.

\subsection{System Level Results}

\begin{table}[t]
\caption{End-to-End 5G Communication Latency by Input Modality. Both results are from real-world testing in a lab environment.}
\label{tab:5g_latency}
\centering
\resizebox{\columnwidth}{!}{
\begin{tabular}{@{}lc|cc@{}}
\toprule
\textbf{Message Type} & \textbf{Payload Size} & \textbf{Public 5G Latency} & \textbf{Private 5G Latency} \\
\midrule
Raw Images        & $\sim$25 MB  & 800--1000 ms          & 430--460 ms \\
BEV Images        & $\sim$4.2 MB & 200--300 ms       & 90--120 ms  \\
Vectorized Messages & $\sim$40 kB  & 80--200 ms        & 20--40 ms   \\
\bottomrule
\end{tabular}
}
\end{table}

\begin{table*}[t]
\centering
\caption{Benchmark comparison on \textit{VecEval} validation set. FleetAgent achieves the lowest intervention failure rate and strong context-aware scores under strict system constraints. \textbf{Bold} denotes the best, and {\ul underlined} denotes the second-best \emph{excluding} the grey Gemini row, which is shown as a reference because it was used to assist annotation during dataset construction.}
\label{tab:model}
\resizebox{\textwidth}{!}{%
\begin{tabular}{cc|cccccc}
\toprule
 &
   &
   &
  \multicolumn{2}{c}{Lingo-Judge} &
  \multicolumn{3}{c}{Language Metrics} \\ \cmidrule(lr){4-5} \cmidrule(lr){6-8}
\multirow{-2}{*}{Model} &
  \multirow{-2}{*}{Input Modality} &
  \multirow{-2}{*}{Intervention Failure Rate (\%)$\downarrow$} &
  Acc. $\uparrow$&
  Score $\uparrow$&
  B $\uparrow$&
  M $\uparrow$&
  R $\uparrow$ \\ \midrule
&
  Raw Images &
  15.24 &
  18.26 &
  0.2304 &
  61.65 &
  24.59 &
  22.34 \\
&
  BEV Images &
  16.05 &
  15.49 &
  0.2191 &
  62.90 &
  24.16 &
  20.92 \\
\multirow{-3}{*}{GPT-4o} &
  Language Description &
  13.63 &
  26.03 &
  0.2422 &
  45.05 &
  29.07 &
  22.21 \\ \midrule
 &
  Raw Images &
  15.90 &
  55.63 &
  0.3281 &
  {\ul 66.28} &
  26.43 &
  27.58 \\
 &
  BEV Images &
  13.83 &
  22.25 &
  0.2533 &
  42.16 &
  21.84 &
  17.78 \\
\multirow{-3}{*}{\begin{tabular}[c]{@{}c@{}}Qwen-2.5VL-7B\\ (Few-shot Example)\end{tabular}} &
  Language Description &
  15.14 &
  52.22 &
  0.3056 &
  48.38 &
  29.84 &
  22.21 \\ \midrule
FleetAgent (w/o tokens prioritization) &
  Vectorized V2N Messages &
  {\ul 12.42} &
  {\ul 54.44} &
  \textbf{0.3599} &
  61.99 &
  \textbf{36.58} &
  \textbf{31.10} \\
FleetAgent &
  Vectorized V2N Messages &
  \textbf{12.12} &
  \textbf{55.93} &
  {\ul 0.3568} &
  \textbf{93.26} &
  {\ul 33.24} &
  {\ul 27.52} \\ \midrule
{\color[HTML]{656565} Gemini-2.5-Flash~(Reference Baseline)} &
  {\color[HTML]{656565} Language Description} &
  {\color[HTML]{656565} 16.00} &
  {\color[HTML]{656565} 78.61} &
  {\color[HTML]{656565} 0.4478} &
  {\color[HTML]{656565} 89.33} &
  {\color[HTML]{656565} 38.86} &
  {\color[HTML]{656565} 37.53} \\ \bottomrule
\end{tabular}%
}
\end{table*}

We compare API-based and local deployment models across four input modalities: raw RGB images, BEV images, tokenized language descriptions, and vectorized V2N messages.
In our setting, language descriptions are deterministically generated from the same structured V2N messages using a fixed prompt template, so packet size matches the vector payload, but the text tokenizer will generate a much longer MLLM context and substantially larger KV-cache than the VecFormer.

As shown in Table~\ref{tab:system}, \method demonstrates superior system efficiency across all aspects. While surround-view raw images and BEV images require massive data transmission ($25,312.5/4,218.8$ kB per request), FleetAgent's vector embeddings deliver similar information with only $40.5$ kB, a $625\times$ reduction in bandwidth requirements. In terms of inference latency, \method achieves the fastest response time of 4.41 seconds, outperforming other methods including locally deployed model Qwen-2.5VL-7B and API service. When compared with language description input, \method requires only $1,241$ MB of cache memory compared to $20,522$ MB for language descriptions (a $16.54\times$ reduction) in local deployments, as the constraints from the text tokenizer are removed and a single vector is encoded into exactly one continuous-space embedding. Though the size of the model is slightly increased with the add-on modules, the significant reduction in cache memory enables faster inference and a larger batch-processing size, given a fixed memory allocation. This provides significant strength in the cloud-hosted teleoperation assistant context, where the data from a massive autonomous fleet is sent to the server simultaneously.

We further test how the transmitted packet size affects communication latency under different real-world 5G conditions.
Table~\ref{tab:5g_latency} shows that although latency does not decrease proportionally with payload size due to connection overhead, vectorized messages retain a clear advantage under both congested public and optimized private networks. Combining transmission latency with model response time yields a substantially lower end-to-end decision latency for \method.

Using vectorized messages during communication and employing \textit{VecFormer} to bridge the input and VLM achieves the most system-level advantages from an architectural perspective.
\subsection{Model Level Results}

\begin{table}[t] 
\caption{Comparison on Nu-X dataset. Baseline results with $^*$ are reported by ALN-P3~\cite{ma_aln-p3_2025}}
\label{tab:nux}
\centering 
\resizebox{\linewidth}{!}{
\begin{tabular}{cc|ccc}
\toprule
\multicolumn{1}{c}{Model} &
  \multicolumn{1}{c}{Input Modality} &
  B &
  M &
  R \\ \midrule
 &
  Raw Images$^*$ &
  3.95 &
  10.3 &
  24.9 \\
\multirow{-2}{*}{GPT-4o} &
  Language Description &
  14.35 &
  11.93 &
  5.85 \\\midrule
 &
  BEV Images &
  31.56 &
  18.63 &
  13.94 \\
\multirow{-2}{*}{Gemini-2.5-Flash} &
  Language Description &
  27.77 &
  18.67 &
  12.75 \\\midrule
 &
  BEV Images &
  23.41 &
  18.25 &
  10.97 \\
\multirow{-2}{*}{\begin{tabular}[c]{@{}l@{}}Qwen-2.5VL-7B\\\end{tabular}} &
  Language Description &
  2.37 &
  8.28 &
  4.29 \\ \midrule
\rowcolor[HTML]{EFEFEF} 
TOD3Cap$^*$ &
  - &
  2.45 &
  10.5 &
  23 \\
\rowcolor[HTML]{EFEFEF} 
HintAD$^*$ &
  - &
  4.18 &
  13.2 &
  27.6 \\
\rowcolor[HTML]{EFEFEF} 
ALN-P3$^*$ &
  - &
  5.59 &
  14.7 &
  35.2 \\ \midrule
FleetAgent &
  Vector Embeddings &
  76.96 &
  19.51 &
  26.72 \\ \bottomrule
\end{tabular}%
}
\vspace{-2em}
\end{table}

The system-level results motivate a structured vector input modality for cloud-based fleet monitoring. Raw and BEV images impose large V2N payloads, whereas textual tokenization of the same structured information produces long MLLM contexts and large KV-cache requirements. The model-level evaluation addresses the complementary feasibility question of whether \textit{VecFormer} can successfully integrate this vector modality into a pretrained image-and-text MLLM while retaining the plan-evaluation and operator-facing explanation capabilities required by the task. Because the general-purpose baselines use different task-adaptation regimes, these cross-model results are interpreted as practical end-to-end references rather than as a controlled comparison.

\begin{figure*}[t]
    \centering
    \includegraphics[width=0.98\linewidth]{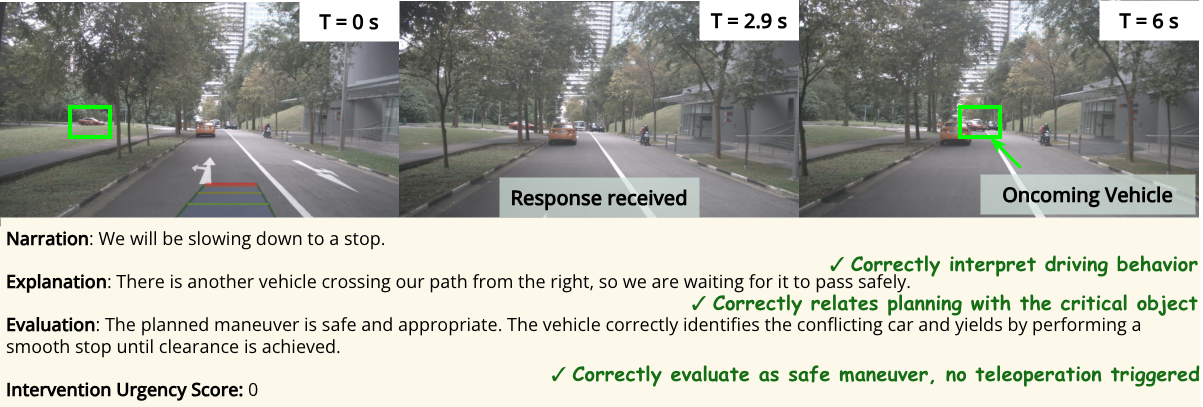}
    \caption{Qualitative example. \method correctly interprets a safe yielding maneuver and assigns low urgency, while some other baselines, including Gemini 2.5-Flash, misinterpret the stop as unsafe.}
    \label{fig:quali}
\end{figure*}

We evaluate models using three metric groups:
(1) \textbf{Context-aware evaluation} via Lingo-Judge \cite{marcu2024lingoqa}, which scores semantic appropriateness and contextual relevance in driving scenarios (we report both raw score and acceptance accuracy with a fixed threshold of 0.3);
(2) \textbf{teleoperation prioritization performance} via Intervention Failure Rate (IFR), defined as the fraction of samples for which the predicted urgency score \(\hat{s}\in[0,10]\) underestimates the human-labeled urgency $s$, or overestimates it by more than a tolerance \(\Delta\):
\[
\mathrm{IFR}= \frac{1}{N}\sum_{i=1}^{N}\mathbf{1}\left[\hat{s}_i < s_i || \hat{s}_i > s_i+\Delta\right],\quad \Delta=2.
\]
We prioritize IFR because, in teleoperation, missing high-urgency cases (underestimating urgency) is typically more costly than conservative overestimation, which mainly increases operator workload;
and (3) \textbf{language metrics} including BLEU~(B),METEOR~(M), and ROUGE~(R), to measure presentation-level similarity and fluency, which are also important in a teleoperator-facing natural language task.

Table~\ref{tab:model} shows that \method achieves the lowest IFR among the tested baselines while maintaining strong context-aware scores under strict system constraints.
Compared with Qwen-2.5VL-7B using language, \method improves Lingo-Judge score by $16.8$\% and reduces intervention failure rate by $19.9$\%, demonstrating that the system-level efficiency gains do not sacrifice plan-evaluation performance.

\noindent\textbf{Ablation on context token prioritization.}
The token prioritization module yields comparable plan-evaluation metrics while reducing the required number of context tokens by 43.7\% and accelerating inference by 14.8\%.

\noindent\textbf{Verification on Nu-X.}
To compare with prior driving reasoning work, we additionally evaluate \method on Nu-X~\cite{ding_hint-ad_2024}. As shown in Table~\ref{tab:nux}, though \method was only trained on the Nu-X dataset while other methods use multiple datasets, including the Nu-X dataset, for extensive VLM grounding, our method can outperform the previous SoTA methods like ALN-P3~\cite{ma_aln-p3_2025}. The results further verify the feasibility of \textit{using highly compact vectorized V2N messages on image and text pretrained VLM.}

\subsection{Qualitative Results}

Fig.~\ref{fig:quali} illustrates that \method can connect the ego plan with the relevant interacting agent and produce an interpretable explanation and evaluation for the operator.
In this scenario, the ego vehicle slows to yield, which is safe and appropriate given the oncoming vehicle.
This example demonstrates that VecFormer can accurately encode the spatial and temporal information of the vehicle-surrounding context and preserve critical interacting elements for reasoning.

\section{Conclusion}
We presented \method, a cloud-hosted MLLM teleoperation assistant that evaluates ego planning using compact vectorized V2N messages and provides structured natural-language explanations with an intervention urgency score for operator prioritization.
We introduced VecFormer to map structured vectors into fixed-length MLLM embeddings with bounded KV-cache growth, and we constructed the VecEval dataset with imperfect plan variants and human-verified vehicle planning evaluation language labels.
Experiments demonstrate substantial system advantages (up to $625\times$ lower uplink payload than raw images and $16.54\times$ lower KV-cache memory than tokenized text descriptions) while maintaining competitive explanation and evaluation performance on VecEval. Due to the inaccessibility of the online teleoperation system, our evaluation is offline, so future work will include closed-loop validation and human-subject studies to assess operator decision-making benefits and to explore adaptive context selection beyond a fixed top-$K$ strategy.



%
\bibliographystyle{IEEEtran}
\bibliography{references,iclr2026_conference}

@inproceedings{marcu2024lingoqa,
  title={Lingoqa: Visual question answering for autonomous driving},
  author={Marcu, Ana-Maria and Chen, Long and H{\"u}nermann, Jan and Karnsund, Alice and Hanotte, Benoit and Chidananda, Prajwal and Nair, Saurabh and Badrinarayanan, Vijay and Kendall, Alex and Shotton, Jamie and others},
  booktitle={European Conference on Computer Vision},
  pages={252--269},
  year={2024},
  organization={Springer}
}

@article{xu_drivegpt4_2024,
    title = {{DriveGPT4}: {Interpretable} {End}-to-{End} {Autonomous} {Driving} {Via} {Large} {Language} {Model}},
    volume = {9},
    issn = {2377-3766},
    shorttitle = {{DriveGPT4}},
    url = {https://ieeexplore.ieee.org/document/10629039/?arnumber=10629039},
    doi = {10.1109/LRA.2024.3440097},
    number = {10},
    urldate = {2024-12-16},
    journal = {IEEE Robotics and Automation Letters},
    author = {Xu, Zhenhua and Zhang, Yujia and Xie, Enze and Zhao, Zhen and Guo, Yong and Wong, Kwan-Yee K. and Li, Zhenguo and Zhao, Hengshuang},
    month = oct,
    year = {2024},
    note = {Conference Name: IEEE Robotics and Automation Letters},
    pages = {8186--8193},
}

@misc{mao_language_2024,
    title = {A {Language} {Agent} for {Autonomous} {Driving}},
    url = {http://arxiv.org/abs/2311.10813},
    doi = {10.48550/arXiv.2311.10813},
    urldate = {2025-09-25},
    publisher = {arXiv},
    author = {Mao, Jiageng and Ye, Junjie and Qian, Yuxi and Pavone, Marco and Wang, Yue},
    month = jul,
    year = {2024},
    note = {arXiv:2311.10813 [cs]},
}

@article{neumeier_data_2022,
    title = {Data {Rate} {Reduction} for {Video} {Streams} in {Teleoperated} {Driving}},
    volume = {23},
    issn = {1558-0016},
    doi = {10.1109/TITS.2022.3171718},
    number = {10},
    journal = {IEEE Transactions on Intelligent Transportation Systems},
    author = {Neumeier, Stefan and Bajpai, Vaibhav and Neumeier, Marion and Facchi, Christian and Ott, Joerg},
    month = oct,
    year = {2022},
    pages = {19145--19160},
}

@inproceedings{schitz_shared_2021,
    title = {Shared {Autonomy} for {Teleoperated} {Driving}: {A} {Real}-{Time} {Interactive} {Path} {Planning} {Approach}},
    shorttitle = {Shared {Autonomy} for {Teleoperated} {Driving}},
    doi = {10.1109/ICRA48506.2021.9561918},
    booktitle = {2021 {IEEE} {International} {Conference} on {Robotics} and {Automation} ({ICRA})},
    author = {Schitz, Dmitrij and Bao, Shuai and Rieth, Dominik and Aschemann, Harald},
    month = may,
    year = {2021},
    note = {ISSN: 2577-087X},
    pages = {999--1004},
}

@inproceedings{georg_sensor_2020,
    title = {Sensor and {Actuator} {Latency} during {Teleoperation} of {Automated} {Vehicles}},
    doi = {10.1109/IV47402.2020.9304802},
    booktitle = {2020 {IEEE} {Intelligent} {Vehicles} {Symposium} ({IV})},
    author = {Georg, Jean-Michael and Feiler, Johannes and Hoffmann, Simon and Diermeyer, Frank},
    month = oct,
    year = {2020},
    note = {ISSN: 2642-7214},
    pages = {760--766},
}

@inproceedings{xie_generative_2021,
    title = {A {Generative} {Model}-{Based} {Predictive} {Display} for {Robotic} {Teleoperation}},
    doi = {10.1109/ICRA48506.2021.9561787},
    booktitle = {2021 {IEEE} {International} {Conference} on {Robotics} and {Automation} ({ICRA})},
    author = {Xie, Bowen and Han, Mingjie and Jin, Jun and Barczyk, Martin and Jägersand, Martin},
    month = may,
    year = {2021},
    note = {ISSN: 2577-087X},
    pages = {2407--2413},
}

@article{gohar_cost_2020,
    title = {A cost efficient multi remote driver selection for remote operated vehicles},
    volume = {168},
    issn = {13891286},
    doi = {10.1016/j.comnet.2019.107029},
    language = {en},
    journal = {Computer Networks},
    author = {Gohar, Ali and Lee, Sanghwan},
    month = feb,
    year = {2020},
    pages = {107029},
}

@misc{wolf_should_2024,
    title = {Should {Teleoperation} {Be} like {Driving} in a {Car}? {Comparison} of {Teleoperation} {HMIs}},
    shorttitle = {Should {Teleoperation} {Be} like {Driving} in a {Car}?},
    url = {http://arxiv.org/abs/2404.13697},
    doi = {10.48550/arXiv.2404.13697},
    urldate = {2025-09-09},
    publisher = {arXiv},
    author = {Wolf, Maria-Magdalena and Taupitz, Richard and Diermeyer, Frank},
    month = apr,
    year = {2024},
    note = {arXiv:2404.13697 [cs]},
}

@misc{tsagkournis_supervised_2023,
    title = {A {Supervised} {Machine} {Learning} {Approach} to {Operator} {Intent} {Recognition} for {Teleoperated} {Mobile} {Robot} {Navigation}},
    url = {http://arxiv.org/abs/2304.14003},
    doi = {10.48550/arXiv.2304.14003},
    urldate = {2025-09-09},
    publisher = {arXiv},
    author = {Tsagkournis, Evangelos and Panagopoulos, Dimitris and Petousakis, Giannis and Nikolaou, Grigoris and Stolkin, Rustam and Chiou, Manolis},
    month = apr,
    year = {2023},
    note = {arXiv:2304.14003 [cs]},
}

@inproceedings{yang_assisted_2020,
    title = {Assisted {Mobile} {Robot} {Teleoperation} with {Intent}-aligned {Trajectories} via {Biased} {Incremental} {Action} {Sampling}},
    doi = {10.1109/IROS45743.2020.9341514},
    booktitle = {2020 {IEEE}/{RSJ} {International} {Conference} on {Intelligent} {Robots} and {Systems} ({IROS})},
    author = {Yang, Xuning and Michael, Nathan},
    month = oct,
    year = {2020},
    note = {ISSN: 2153-0866},
    pages = {10998--11003},
}

@article{cho_goondae_2023,
    title = {{GoonDAE}: {Denoising}-{Based} {Driver} {Assistance} for {Off}-{Road} {Teleoperation}},
    volume = {8},
    issn = {2377-3766, 2377-3774},
    shorttitle = {{GoonDAE}},
    doi = {10.1109/LRA.2023.3250008},
    number = {4},
    journal = {IEEE Robotics and Automation Letters},
    author = {Cho, Younggeol and Yun, Hyeonggeun and Lee, Jinwon and Ha, Arim and Yun, Jihyeok},
    month = apr,
    year = {2023},
    note = {arXiv:2209.03568 [cs]},
    pages = {2405--2412},
}

@misc{rahmani_systematic_2024,
    title = {A {Systematic} {Review} of {Edge} {Case} {Detection} in {Automated} {Driving}: {Methods}, {Challenges} and {Future} {Directions}},
    shorttitle = {A {Systematic} {Review} of {Edge} {Case} {Detection} in {Automated} {Driving}},
    url = {http://arxiv.org/abs/2410.08491},
    doi = {10.48550/arXiv.2410.08491},
    urldate = {2025-09-09},
    publisher = {arXiv},
    author = {Rahmani, Saeed and Rieder, Sabine and Gelder, Erwin de and Sonntag, Marcel and Mallada, Jorge Lorente and Kalisvaart, Sytze and Hashemi, Vahid and Calvert, Simeon C.},
    month = oct,
    year = {2024},
    note = {arXiv:2410.08491 [cs]},
}

@misc{wang_generative_2025,
    title = {Generative {AI} for {Autonomous} {Driving}: {Frontiers} and {Opportunities}},
    shorttitle = {Generative {AI} for {Autonomous} {Driving}},
    url = {http://arxiv.org/abs/2505.08854},
    doi = {10.48550/arXiv.2505.08854},
    urldate = {2025-09-10},
    publisher = {arXiv},
    author = {Wang, Yuping and Xing, Shuo and Can, Cui and Li, Renjie and Hua, Hongyuan and Tian, Kexin and Mo, Zhaobin and Gao, Xiangbo and Wu, Keshu and Zhou, Sulong and You, Hengxu and Peng, Juntong and Zhang, Junge and Wang, Zehao and Song, Rui and Yan, Mingxuan and Zimmer, Walter and Zhou, Xingcheng and Li, Peiran and Lu, Zhaohan and Chen, Chia-Ju and Huang, Yue and Rossi, Ryan A. and Sun, Lichao and Yu, Hongkai and Fan, Zhiwen and Yang, Frank Hao and Kang, Yuhao and Greer, Ross and Liu, Chenxi and Lee, Eun Hak and Di, Xuan and Ye, Xinyue and Ren, Liu and Knoll, Alois and Li, Xiaopeng and Ji, Shuiwang and Tomizuka, Masayoshi and Pavone, Marco and Yang, Tianbao and Du, Jing and Yang, Ming-Hsuan and Wei, Hua and Wang, Ziran and Zhou, Yang and Li, Jiachen and Tu, Zhengzhong},
    month = may,
    year = {2025},
    note = {arXiv:2505.08854 [cs]},
}

@inproceedings{sima_drivelm_2025,
    address = {Cham},
    title = {{DriveLM}: {Driving} with {Graph} {Visual} {Question} {Answering}},
    isbn = {978-3-031-72943-0},
    shorttitle = {{DriveLM}},
    doi = {10.1007/978-3-031-72943-0_15},
    language = {en},
    booktitle = {Computer {Vision} – {ECCV} 2024},
    publisher = {Springer Nature Switzerland},
    author = {Sima, Chonghao and Renz, Katrin and Chitta, Kashyap and Chen, Li and Zhang, Hanxue and Xie, Chengen and Beißwenger, Jens and Luo, Ping and Geiger, Andreas and Li, Hongyang},
    editor = {Leonardis, Aleš and Ricci, Elisa and Roth, Stefan and Russakovsky, Olga and Sattler, Torsten and Varol, Gül},
    year = {2025},
    pages = {256--274},
}

@misc{tian_drivevlm_2024,
    title = {{DriveVLM}: {The} {Convergence} of {Autonomous} {Driving} and {Large} {Vision}-{Language} {Models}},
    shorttitle = {{DriveVLM}},
    url = {http://arxiv.org/abs/2402.12289},
    doi = {10.48550/arXiv.2402.12289},
    urldate = {2025-02-20},
    publisher = {arXiv},
    author = {Tian, Xiaoyu and Gu, Junru and Li, Bailin and Liu, Yicheng and Wang, Yang and Zhao, Zhiyong and Zhan, Kun and Jia, Peng and Lang, Xianpeng and Zhao, Hang},
    month = jun,
    year = {2024},
    note = {arXiv:2402.12289 [cs]},
}

@misc{park_nuplanqa_2025,
    title = {{NuPlanQA}: {A} {Large}-{Scale} {Dataset} and {Benchmark} for {Multi}-{View} {Driving} {Scene} {Understanding} in {Multi}-{Modal} {Large} {Language} {Models}},
    shorttitle = {{NuPlanQA}},
    url = {http://arxiv.org/abs/2503.12772},
    doi = {10.48550/arXiv.2503.12772},
    urldate = {2025-09-10},
    publisher = {arXiv},
    author = {Park, Sung-Yeon and Cui, Can and Ma, Yunsheng and Moradipari, Ahmadreza and Gupta, Rohit and Han, Kyungtae and Wang, Ziran},
    month = aug,
    year = {2025},
    note = {arXiv:2503.12772 [cs]},
}

@misc{li_recogdrive_2025,
    title = {{ReCogDrive}: {A} {Reinforced} {Cognitive} {Framework} for {End}-to-{End} {Autonomous} {Driving}},
    shorttitle = {{ReCogDrive}},
    url = {http://arxiv.org/abs/2506.08052},
    doi = {10.48550/arXiv.2506.08052},
    urldate = {2025-09-10},
    publisher = {arXiv},
    author = {Li, Yongkang and Xiong, Kaixin and Guo, Xiangyu and Li, Fang and Yan, Sixu and Xu, Gangwei and Zhou, Lijun and Chen, Long and Sun, Haiyang and Wang, Bing and Chen, Guang and Ye, Hangjun and Liu, Wenyu and Wang, Xinggang},
    month = jun,
    year = {2025},
    note = {arXiv:2506.08052 [cs]},
}

@misc{hwang_emma_2024,
    title = {{EMMA}: {End}-to-{End} {Multimodal} {Model} for {Autonomous} {Driving}},
    shorttitle = {{EMMA}},
    url = {http://arxiv.org/abs/2410.23262},
    doi = {10.48550/arXiv.2410.23262},
    urldate = {2025-05-09},
    publisher = {arXiv},
    author = {Hwang, Jyh-Jing and Xu, Runsheng and Lin, Hubert and Hung, Wei-Chih and Ji, Jingwei and Choi, Kristy and Huang, Di and He, Tong and Covington, Paul and Sapp, Benjamin and Guo, James and Anguelov, Dragomir and Tan, Mingxing},
    month = oct,
    year = {2024},
    note = {arXiv:2410.23262 [cs]
version: 1},
}

@misc{zhou_opendrivevla_2025,
    title = {{OpenDriveVLA}: {Towards} {End}-to-end {Autonomous} {Driving} with {Large} {Vision} {Language} {Action} {Model}},
    shorttitle = {{OpenDriveVLA}},
    url = {http://arxiv.org/abs/2503.23463},
    doi = {10.48550/arXiv.2503.23463},
    urldate = {2025-09-10},
    publisher = {arXiv},
    author = {Zhou, Xingcheng and Han, Xuyuan and Yang, Feng and Ma, Yunpu and Knoll, Alois C.},
    month = mar,
    year = {2025},
    note = {arXiv:2503.23463 [cs]},
}

@misc{ding_hint-ad_2024,
    title = {Hint-{AD}: {Holistically} {Aligned} {Interpretability} in {End}-to-{End} {Autonomous} {Driving}},
    shorttitle = {Hint-{AD}},
    url = {http://arxiv.org/abs/2409.06702},
    doi = {10.48550/arXiv.2409.06702},
    urldate = {2025-06-05},
    publisher = {arXiv},
    author = {Ding, Kairui and Chen, Boyuan and Su, Yuchen and Gao, Huan-ang and Jin, Bu and Sima, Chonghao and Zhang, Wuqiang and Li, Xiaohui and Barsch, Paul and Li, Hongyang and Zhao, Hao},
    month = sep,
    year = {2024},
    note = {arXiv:2409.06702 [cs]},
}

@misc{ma_aln-p3_2025,
    title = {{ALN}-{P3}: {Unified} {Language} {Alignment} for {Perception}, {Prediction}, and {Planning} in {Autonomous} {Driving}},
    shorttitle = {{ALN}-{P3}},
    url = {http://arxiv.org/abs/2505.15158},
    doi = {10.48550/arXiv.2505.15158},
    urldate = {2025-09-10},
    publisher = {arXiv},
    author = {Ma, Yunsheng and Yaman, Burhaneddin and Ye, Xin and Yurt, Mahmut and Luo, Jingru and Mallik, Abhirup and Wang, Ziran and Ren, Liu},
    month = may,
    year = {2025},
    note = {arXiv:2505.15158 [cs]},
}

@misc{testouri_5g-enabled_2025,
    title = {{5G}-{Enabled} {Teleoperated} {Driving}: {An} {Experimental} {Evaluation}},
    shorttitle = {{5G}-{Enabled} {Teleoperated} {Driving}},
    url = {http://arxiv.org/abs/2503.14186},
    doi = {10.48550/arXiv.2503.14186},
    urldate = {2025-09-10},
    publisher = {arXiv},
    author = {Testouri, Mehdi and Elghazaly, Gamal and Hawlader, Faisal and Frank, Raphael},
    month = mar,
    year = {2025},
    note = {arXiv:2503.14186 [cs]},
}

@misc{zhao_quantv2x_2025,
    title = {{QuantV2X}: {A} {Fully} {Quantized} {Multi}-{Agent} {System} for {Cooperative} {Perception}},
    shorttitle = {{QuantV2X}},
    url = {http://arxiv.org/abs/2509.03704},
    doi = {10.48550/arXiv.2509.03704},
    urldate = {2025-09-10},
    publisher = {arXiv},
    author = {Zhao, Seth Z. and Zhang, Huizhi and Li, Zhaowei and Peng, Juntong and Chui, Anthony and Zhou, Zewei and Meng, Zonglin and Xiang, Hao and Huang, Zhiyu and Wang, Fujia and Tian, Ran and Xu, Chenfeng and Zhou, Bolei and Ma, Jiaqi},
    month = sep,
    year = {2025},
    note = {arXiv:2509.03704 [cs]},
}

@inproceedings{caesar_nuscenes_2020,
    title = {{nuScenes}: {A} {Multimodal} {Dataset} for {Autonomous} {Driving}},
    shorttitle = {{nuScenes}},
    urldate = {2025-05-09},
    author = {Caesar, Holger and Bankiti, Varun and Lang, Alex H. and Vora, Sourabh and Liong, Venice Erin and Xu, Qiang and Krishnan, Anush and Pan, Yu and Baldan, Giancarlo and Beijbom, Oscar},
    year = {2020},
    pages = {11621--11631},
}

@inproceedings{sun_scalability_2020,
    title = {Scalability in {Perception} for {Autonomous} {Driving}: {Waymo} {Open} {Dataset}},
    shorttitle = {Scalability in {Perception} for {Autonomous} {Driving}},
    urldate = {2025-05-09},
    author = {Sun, Pei and Kretzschmar, Henrik and Dotiwalla, Xerxes and Chouard, Aurelien and Patnaik, Vijaysai and Tsui, Paul and Guo, James and Zhou, Yin and Chai, Yuning and Caine, Benjamin and Vasudevan, Vijay and Han, Wei and Ngiam, Jiquan and Zhao, Hang and Timofeev, Aleksei and Ettinger, Scott and Krivokon, Maxim and Gao, Amy and Joshi, Aditya and Zhang, Yu and Shlens, Jonathon and Chen, Zhifeng and Anguelov, Dragomir},
    year = {2020},
    pages = {2446--2454},
}

@article{geiger_vision_2013,
    title = {Vision meets robotics: {The} {KITTI} dataset},
    volume = {32},
    issn = {0278-3649, 1741-3176},
    shorttitle = {Vision meets robotics},
    doi = {10.1177/0278364913491297},
    language = {en},
    number = {11},
    journal = {The International Journal of Robotics Research},
    author = {Geiger, A and Lenz, P and Stiller, C and Urtasun, R},
    month = sep,
    year = {2013},
    pages = {1231--1237},
}

@misc{caesar_nuplan_2022,
    title = {{NuPlan}: {A} closed-loop {ML}-based planning benchmark for autonomous vehicles},
    shorttitle = {{NuPlan}},
    url = {http://arxiv.org/abs/2106.11810},
    doi = {10.48550/arXiv.2106.11810},
    urldate = {2025-05-09},
    publisher = {arXiv},
    author = {Caesar, Holger and Kabzan, Juraj and Tan, Kok Seang and Fong, Whye Kit and Wolff, Eric and Lang, Alex and Fletcher, Luke and Beijbom, Oscar and Omari, Sammy},
    month = feb,
    year = {2022},
    note = {arXiv:2106.11810 [cs]},
}

@misc{qian_nuscenes-qa_2024,
    title = {{NuScenes}-{QA}: {A} {Multi}-modal {Visual} {Question} {Answering} {Benchmark} for {Autonomous} {Driving} {Scenario}},
    shorttitle = {{NuScenes}-{QA}},
    url = {http://arxiv.org/abs/2305.14836},
    doi = {10.48550/arXiv.2305.14836},
    urldate = {2025-09-12},
    publisher = {arXiv},
    author = {Qian, Tianwen and Chen, Jingjing and Zhuo, Linhai and Jiao, Yang and Jiang, Yu-Gang},
    month = feb,
    year = {2024},
    note = {arXiv:2305.14836 [cs]},
}

@article{wu_language_2025,
    title = {Language {Prompt} for {Autonomous} {Driving}},
    volume = {39},
    copyright = {Copyright (c) 2025 Association for the Advancement of Artificial Intelligence},
    issn = {2374-3468},
    doi = {10.1609/aaai.v39i8.32902},
    language = {en},
    number = {8},
    journal = {Proceedings of the AAAI Conference on Artificial Intelligence},
    author = {Wu, Dongming and Han, Wencheng and Liu, Yingfei and Wang, Tiancai and Xu, Cheng-Zhong and Zhang, Xiangyu and Shen, Jianbing},
    month = apr,
    year = {2025},
    pages = {8359--8367},
}

@InProceedings{Ding_2024_CVPR,
    author    = {Ding, Xinpeng and Han, Jianhua and Xu, Hang and Liang, Xiaodan and Zhang, Wei and Li, Xiaomeng},
    title     = {Holistic Autonomous Driving Understanding by Bird's-Eye-View Injected Multi-Modal Large Models},
    booktitle = {Proceedings of the IEEE/CVF Conference on Computer Vision and Pattern Recognition (CVPR)},
    month     = {June},
    year      = {2024},
    pages     = {13668-13677}
}

@incollection{ferrari_textual_2018,
    address = {Cham},
    title = {Textual {Explanations} for {Self}-{Driving} {Vehicles}},
    volume = {11206},
    language = {en},
    booktitle = {Computer {Vision} – {ECCV} 2018},
    publisher = {Springer International Publishing},
    author = {Kim, Jinkyu and Rohrbach, Anna and Darrell, Trevor and Canny, John and Akata, Zeynep},
    editor = {Ferrari, Vittorio and Hebert, Martial and Sminchisescu, Cristian and Weiss, Yair},
    year = {2018},
    doi = {10.1007/978-3-030-01216-8_35},
    note = {Series Title: Lecture Notes in Computer Science},
    pages = {577--593},
}

@misc{bai_qwen25-vl_2025,
    title = {Qwen2.5-{VL} {Technical} {Report}},
    url = {http://arxiv.org/abs/2502.13923},
    doi = {10.48550/arXiv.2502.13923},
    urldate = {2025-09-19},
    publisher = {arXiv},
    author = {Bai, Shuai and Chen, Keqin and Liu, Xuejing and Wang, Jialin and Ge, Wenbin and Song, Sibo and Dang, Kai and Wang, Peng and Wang, Shijie and Tang, Jun and Zhong, Humen and Zhu, Yuanzhi and Yang, Mingkun and Li, Zhaohai and Wan, Jianqiang and Wang, Pengfei and Ding, Wei and Fu, Zheren and Xu, Yiheng and Ye, Jiabo and Zhang, Xi and Xie, Tianbao and Cheng, Zesen and Zhang, Hang and Yang, Zhibo and Xu, Haiyang and Lin, Junyang},
    month = feb,
    year = {2025},
    note = {arXiv:2502.13923 [cs]},
}

@misc{jang_categorical_2017,
    title = {Categorical {Reparameterization} with {Gumbel}-{Softmax}},
    url = {http://arxiv.org/abs/1611.01144},
    doi = {10.48550/arXiv.1611.01144},
    urldate = {2025-09-23},
    publisher = {arXiv},
    author = {Jang, Eric and Gu, Shixiang and Poole, Ben},
    month = aug,
    year = {2017},
    note = {arXiv:1611.01144 [stat]},
}

@misc{waymo,
  author       = {{Waymo}},
  title        = {Waymo},
  year         = {2025},
  howpublished = {\url{https://waymo.com/}},
  note         = {Accessed: 2025-09-24}
}

@misc{zoox,
  author       = {{Zoox}},
  title        = {Zoox: It's Not a Car},
  year         = {2025},
  howpublished = {\url{https://zoox.com/}},
  note         = {Accessed: 2025-09-24}
}

@misc{tesla_robotaxi,
  author       = {{Tesla}},
  title        = {Tesla Robotaxi},
  year         = {2025},
  howpublished = {\url{https://www.tesla.com/robotaxi}},
  note         = {Accessed: 2025-09-23}
}

@article{abboud2016interworking,
  title={Interworking of DSRC and cellular network technologies for V2X communications: A survey},
  author={Abboud, Khadige and Omar, Hassan Aboubakr and Zhuang, Weihua},
  journal={IEEE transactions on vehicular technology},
  volume={65},
  number={12},
  pages={9457--9470},
  year={2016},
  publisher={IEEE}
}

\end{document}